\documentclass[conference]{IEEEtran}
\IEEEoverridecommandlockouts

\usepackage{dsfont}
\usepackage{multirow}
\usepackage[ruled,vlined]{algorithm2e}
\usepackage{algpseudocode}
\usepackage{tabularx}

\usepackage{cite}
\usepackage{amsmath,amssymb,amsfonts}
\usepackage{graphicx}
\usepackage{textcomp}
\usepackage{xcolor}
\def\BibTeX{{\rm B\kern-.05em{\sc i\kern-.025em b}\kern-.08em
    T\kern-.1667em\lower.7ex\hbox{E}\kern-.125emX}}
\begin{document}

\title{Contrastive Token-level Explanations for Graph-based Rumour Detection}

\author{\IEEEauthorblockN{Daniel Wai Kit Chin}
\IEEEauthorblockA{\textit{Information Systems Technology and Design Pillar} \\
\textit{Singapore University of Technology and Design}\\
Singapore, Singapore \\
daniel\_chin@mymail.sutd.edu.sg}
\and
\IEEEauthorblockN{Roy Ka-Wei Lee}
\IEEEauthorblockA{\textit{Information Systems Technology and Design Pillar} \\
\textit{Singapore University of Technology and Design}\\
Singapore, Singapore \\
roy\_lee@sutd.edu.sg}
}

\maketitle

\begin{abstract}
The widespread use of social media has accelerated the dissemination of information, but it has also facilitated the spread of harmful rumours, which can disrupt economies, influence political outcomes, and exacerbate public health crises, such as the COVID-19 pandemic. While Graph Neural Network (GNN)-based approaches have shown significant promise in automated rumour detection, they often lack transparency, making their predictions difficult to interpret. Existing graph explainability techniques fall short in addressing the unique challenges posed by the dependencies among feature dimensions in high-dimensional text embeddings used in GNN-based models. In this paper, we introduce Contrastive Token Layerwise Relevance Propagation (CT-LRP), a novel framework designed to enhance the explainability of GNN-based rumour detection. CT-LRP extends current graph explainability methods by providing token-level explanations that offer greater granularity and interpretability. We evaluate the effectiveness of CT-LRP across multiple GNN models trained on three publicly available rumour detection datasets, demonstrating that it consistently produces high-fidelity, meaningful explanations, paving the way for more robust and trustworthy rumour detection systems.
\end{abstract}

\begin{IEEEkeywords}
rumour detection, graph neural network, explainability, interpretability
\end{IEEEkeywords}

\section{Introduction}
\label{sec:intro}
\textbf{Motivation.} Social media platforms have revolutionized communication, enabling rapid information sharing but also amplifying the spread of misinformation, including rumours and fake news \cite{vosoughi2018spread}. Crises like the the Russia-Ukraine war highlight the susceptibility of users to such content \cite{aimeur2023fake}. The unchecked dissemination of rumours can cause significant harm \cite{ahsan2019rumors}, emphasizing the need for automated detection methods that mitigate the spread of rumours \cite{thorne2018automated}. To meet this challenge, it is essential to develop trustworthy tools that not only detect rumours effectively but also provide clear, interpretable explanations for their predictions.

Early rumour detection methods relied on text mining and handcrafted features \cite{castillo2011information, yang2012automatic, liu2015real}. While these approaches laid a foundation, their reliance on manually engineered features limited scalability. 
Deep learning methods, such as Recurrent Neural Networks (RNNs) \cite{ma2016detecting} and Long Short-Term Memory (LSTM) networks \cite{kochkina-etal-2017-turing}, improved detection by capturing temporal dependencies in rumour propagation.
However, these models fail to incorporate the structural information unique to rumours, prompting the development of approaches that leverage propagation structures through kernel models \cite{ma2017detect}, Recursive Neural Networks (RvNN) \cite{ma2018rumor}, and Graph Neural Networks (GNN) \cite{Bian2020RumorDO, wei-etal-2021-towards, lin-etal-2021-rumor}. GNNs, in particular, have demonstrated strong performance and computational efficiency, making them effective for both rumour detection and broader misinformation challenges \cite{phan2023fake, guo2022survey}.

Graph explainability techniques, widely used in domains such as molecular chemistry \cite{reiser2022graph, li2021graph}, citation networks \cite{xiao2022graph, chunaev2020community}, and scene graphs \cite{chang2021comprehensive, zareian2020bridging}, have seen limited application in misinformation detection. For GNN-based rumor detection, enhancing explainability is crucial for improving model trust and reliability. Techniques can be broadly categorized as gradient-based, decomposition-based, perturbation-based, and surrogate-based \cite{yuan2022explainability}. While perturbation-based and surrogate-based methods provide powerful insights, they are computationally intensive and lack generalizability in dynamic rumor contexts \cite{yuan2022explainability}. In contrast, gradient-based and decomposition-based approaches offer efficiency and scalability by leveraging model internal mechanisms.

However, current explainability methods often provide only node or edge-level insights, which fail to capture critical dependencies in high-dimensional text embeddings used as node features in GNN-based models \cite{Bian2020RumorDO, wei-etal-2021-towards, lin-etal-2021-rumor}. To address this limitation, explanations must go beyond coarse representations to consider individual textual components, offering finer granularity and higher fidelity.

\textbf{Research Objectives.} We propose Contrastive Token Layerwise Relevance Propagation (CT-LRP), a novel framework that addresses the limitations of existing GNN explainability techniques by providing fine-grained, token-level explanations for rumour detection models. CT-LRP combines Layerwise Relevance Propagation (LRP) with an explanation space partitioning strategy, enabling it to isolate class-specific and task-relevant textual components. This token-level granularity captures dependencies in high-dimensional text embeddings, offering nuanced insights into model predictions that surpass traditional node and edge-level explanations.

To rigorously evaluate CT-LRP, we extend existing explanation metrics to support token-level resolution, ensuring fidelity and interpretability. Experiments on three public rumour detection datasets demonstrate that CT-LRP consistently produces reliable, high-quality explanations, setting a new standard for GNN-based explainability in misinformation detection. By enhancing transparency and trustworthiness, CT-LRP directly addresses the societal need for ethical and effective AI systems to combat misinformation and its harmful consequences.

To summarize, the main contributions of our work are: \begin{itemize} 
    \item We introduce CT-LRP, a post hoc framework that provides granular, high-fidelity token-level explanations for GNN-based rumour detection models.
    \item We extend existing evaluation metrics to support token-level resolution, enabling robust and accurate assessments of explanations. 
    \item We validate CT-LRP through experiments on three public datasets, demonstrating its effectiveness in producing reliable and interpretable explanations.
\end{itemize}

\textbf{Boarder Impact.} Our work advances the technical state-of-the-art explainability for GNN-based rumour detection and addresses the broader societal challenge of combating misinformation. By providing interpretable, token-level explanations, CT-LRP enhances transparency and trust in AI systems, empowering stakeholders to make informed decisions. This framework represents a step toward building ethical and accountable AI tools that mitigate the harm caused by misinformation, fostering a safer online space.



\section{Related Work}
\label{sec:related}
In this section, we discuss the relation of our work to existing research in GNN explainability and text explainability.

\subsection{GNN Explainability}
Explainability techniques for GNNs can be broadly categorized into two groups: (i) methods adapted from Convolutional Neural Networks (CNNs) and (ii) methods specifically designed for GNNs. The first category includes gradient-based and decomposition-based methods such as LRP \cite{bach2015pixel}, Grad-CAM \cite{Selvaraju2017ICCV}, Excitation Backpropagation (EB) \cite{zhang2018top}, and Sensitivity Analysis (SA) \cite{gevrey2003review}, as well as perturbation-based approaches like LIME \cite{ribeiro2016should} and SHAP \cite{lundberg2017unified}. These techniques, originally developed for CNNs, generalize well to GNNs by treating graphs as lattice-shaped structures, where nodes represent pixels or features, and convolution filters act as subgraph kernels. However, these methods typically treat nodes like pixels with node features analogous to colour channels which fail to capture the nuance within the text represented by the node.

The second category includes techniques specifically designed for GNNs, such as GNNExplainer \cite{ying2019gnnexplainer}, PGExplainer \cite{luo2020parameterized}, ZORRO \cite{funke2022zorro}, and GraphLIME \cite{huang2022graphlime}. These methods provide insights into node, edge, or subgraph-level attributions, offering explainability for graph-specific tasks like node classification or link prediction. However, their reliance on perturbation or surrogate models often limits their generalizability to unseen data, particularly in dynamic applications like rumour detection. Gradient-based methods, while more scalable, lack the granularity needed to interpret latent textual features represented in high-dimensional node embeddings.

Our framework, CT-LRP, addresses these limitations by extending GNN explainability to token-level resolution, overcoming the challenges of interpreting latent text features in rumor detection. By leveraging the strengths of gradient-based and decomposition-based methods, CT-LRP provides fine-grained insights that generalize effectively to unseen data while maintaining high fidelity.

\subsection{Text Explainability}
Explainability methods originally developed for text classification tasks can be categorized into gradient-based, decomposition-based, and perturbation-based approaches. Gradient-based and decomposition-based methods, including LRP \cite{bach2015pixel}, Saliency Maps \cite{simonyan2013deep}, and Guided Backpropagation \cite{springenberg2014striving}, trace backpropagated gradients or relevance scores to determine the contribution of individual tokens to the model's decision. Perturbation-based techniques such as LIME \cite{ribeiro2016should}, SHAP \cite{lundberg2017unified}, and Occlusion \cite{zeiler2014visualizing} identify influential tokens by substituting input elements and observing changes in model output. These methods often struggle with latent feature interpretability, particularly when applied to high-dimensional embeddings.

Our approach recasts the explainability problem for GNN-based rumor detection as a text explanation task. By combining GNN explainability methods to accurately attribute node features with text explainability techniques to generate token-level explanations, CT-LRP bridges the gap between node feature attribution and interpretable token-level insights. This hybrid approach ensures that latent text features are effectively explained, offering a novel solution to the challenges faced by existing methods.

\begin{table}[t]
\caption{Definition of Math Notation and Symbols in this Paper}
\centering
\begin{tabular}{|l||l|}
\hline
$h, H, \textbf{h}, \textbf{H}$ & scalar, set, vector, matrix \\
$\textbf{h}_i, \textbf{H}_{i,j}$ & vector entry, matrix entry \\
\hline
$G, V, E$ & graph, vertices, edges \\ 
$\textbf{A}$ & adjacency matrix\\
$\textbf{y}, c, C$ & output vector, output class, set of output classes \\
$\textbf{x}, \textbf{X}, d, D$ & feature vector, feature matrix, feature, feature set \\
$P, T, T_v$ & tokenised text set, token set, token set for node $v$ \\
$f, \Theta$ & function, function parameter set \\
$LRP$ & Layerwise Relevance Propagation \\
$\textbf{Z}, \textbf{R}$ & token relevance, node relevance \\
$\textbf{Z}^{(c)}, \textbf{R}^{(c)}$ & token and node relevance w.r.t class  $c$ \\
$v, t$ & integers for identifying vertices and tokens \\
\hline
\end{tabular}
\label{tab:notation}
\end{table}

\begin{figure*}[hbt!]
    \centering
    \includegraphics[width=\linewidth]{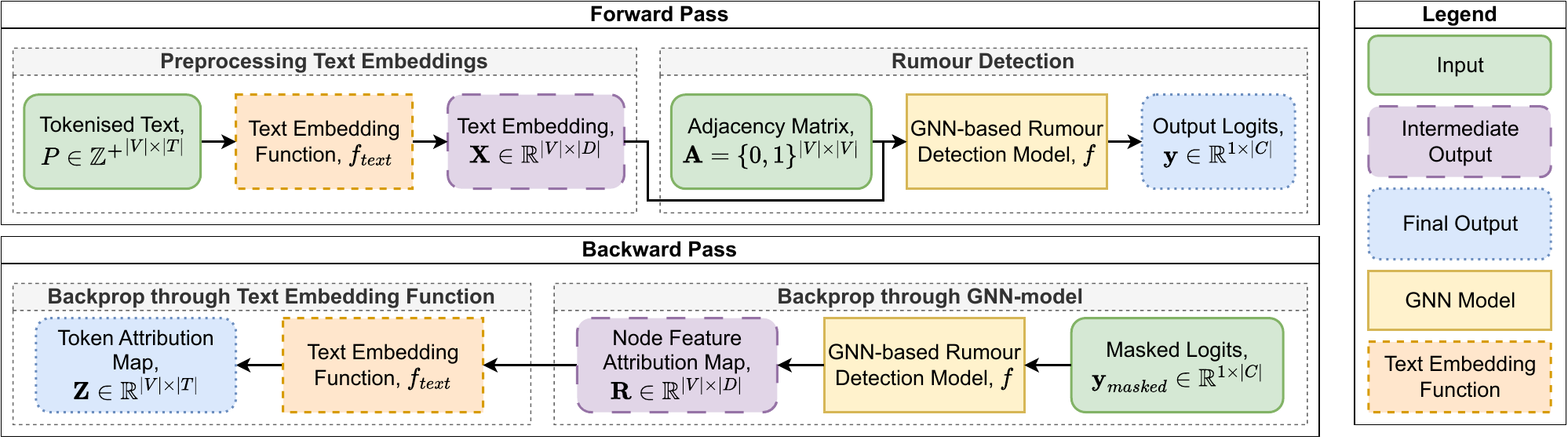}
    \caption{Overview of the proposed CT-LRP framework showing the flow of information through the GNN and text embedding function. Inputs to the forward and backward pass are colour-coded in green, intermediate outputs in purple and final outputs in blue.}
    \label{fig:ct-lrp}
\end{figure*}

\section{Preliminaries}
\label{sec:preliminaries}



In rumour detection, event propagation on social media is modelled as a \textbf{graph-level classification task}. Let \( G = (V, E) \) represent an event propagation graph, where \( V \) is the set of nodes corresponding to posts made during the event, and \( E \) is the set of edges capturing interactions between these posts. Each node \( v \in V \) is associated with a feature vector \( \mathbf{x}_v \in \mathbb{R}^{|D|} \), typically a text embedding derived from the post content. The feature matrix \( \mathbf{X} \in \mathbb{R}^{|V| \times |D|} \) aggregates the embeddings for all posts in the event. The notations used are summarized in Table \ref{tab:notation}.

The goal of the task is to learn a function \( f \) that given an information propagation event graph \( G \) predicts whether the source post \( v_0 \) of that graph is a rumour, i.e., \( f(G) = \hat{y} \). 
This can be formulated as either a binary classification problem, as in the Weibo dataset~\cite{ma2016detecting}, or a multi-class classification problem, as in the Twitter15/16 datasets~\cite{ma2016detecting}. In this paper, we adopt GNN-based models as \( f \), leveraging their ability to capture both textual and structural information within the event graph.

Some approaches extend this task by incorporating additional features, such as user attributes or handcrafted graph features (e.g., node centrality), into the graph structure. For example, user attributes like follower count or account age can be represented as nodes in a user graph \( G_{\text{user}} \). While these features can provide additional context, they rely on explicit, handcrafted representations that are straightforward to interpret with existing explainability methods. In contrast, our focus lies in enhancing the interpretability of GNN-based models operating on latent, high-dimensional text embeddings, which pose greater challenges for explainability.

By improving the explainability of these latent representations, we aim to generate fine-grained insights into model predictions, addressing a critical gap in existing methods for rumour detection.

\section{Proposed Framework}
\label{sec:framework}


\begin{figure*}[hbt!]
    \centering
    \includegraphics[width=\linewidth]{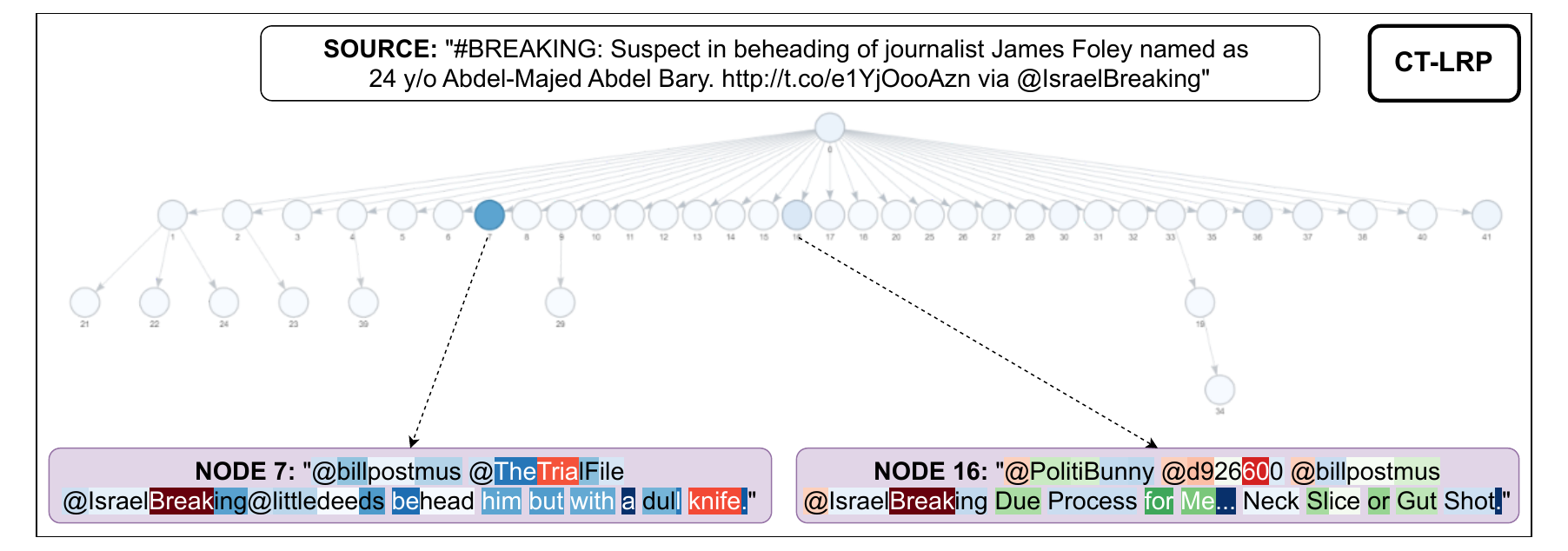}
    \caption{Token-level Explanation produced by CT-LRP. Node-level attribution is shown with darker shades of blue indicating greater attribution from that node. Token-level explanation with blue and green highlights for class-specific and common task-relevant attribution, and red highlights for negative attribution. Darker shades indicate a greater magnitude of importance. The input is the source claim and the responses are modelled as a graph.}
    \label{fig:ct-lrp-vis}
\end{figure*}

Our approach addresses the low resolution of existing GNN explainability methods by reframing the explanation process to align with techniques used in text classification models. Instead of attributing relevance to entire sentences or posts represented as nodes in the graph, we aim to pinpoint the specific tokens that drive the model's predictions for a given class. This fine-grained approach enhances both the granularity and interpretability of the explanations. In this subsection, we first formalize the task of explaining GNN-based rumour detection models within the context of event propagation graphs. We then introduce our framework, which delivers detailed attributions at both the GNN-encoder and token levels, bridging the gap between high-dimensional embeddings and actionable insights.

\subsection{Task Formulation}
Building on the graph representation outlined in Section \ref{sec:preliminaries}, we expand the attributed event propagation graph \( G = (V, E, \mathbf{X}) \) to incorporate tokenized text content \( P \) associated with the nodes. Here, \( V = \{0, 1, \ldots, n-1\} \) is the set of nodes, \( E \) is the set of edges, and \( \mathbf{X} \in \mathbb{R}^{|V| \times |D|} \) is the node feature matrix. The tokenized text \( P \in \mathbb{Z^+}^{|V| \times |T|} \) is mapped to the feature matrix \( \mathbf{X} \) via a text embedding function \( f_{\text{text}}(P | \Theta_{\text{text}}) = \mathbf{X} \), where \( T \) represents the set of tokens for each node. The rumour detection model is implemented as a GNN \( f(G | \Theta) = \mathbf{y} \), parameterized by \( \Theta \). Given \( G \), the model outputs logits \( \mathbf{y} \in \mathbb{R}^{|C|} \), where the predicted class \( \hat{y} \) is obtained as \( \hat{y} = \text{argmax}(\mathbf{y}) \). The goal of explainability is to generate a token-level attribution map \( \mathbf{Z}^{(\hat{y})} = [\mathbf{z}_0^\text{T}, \ldots, \mathbf{z}_v^\text{T}, \ldots, \mathbf{z}_{n-1}^\text{T}]^\text{T} \), where \( \mathbf{z}_v \in \mathbb{R}^{|T_v|} \) quantifies the contribution of each token \( t \in T_v \) in node \( v \) to the prediction \( \hat{y} \).

\textbf{Framework Overview.} Fig.~\ref{fig:ct-lrp} illustrates the flow of information through our framework. The process involves a \textit{forward pass} and a \textit{backward pass}. In the \textit{Forward Pass}, tokenized text \( P \) is transformed into the feature matrix \( \mathbf{X} \) via \( f_{\text{text}} \). The GNN processes \( \mathbf{X} \) along with the adjacency matrix \( \mathbf{A} \in \{0, 1\}^{|V| \times |V|} \), producing logits \( \mathbf{y} \). In the \textit{Backward Pass}, relevance scores are calculated for the predicted class \( \hat{y} \) by masking all other logits. These scores are backpropagated through the GNN to generate node-level attributions \( \mathbf{R}_{\text{node}} \). Finally, \( \mathbf{R}_{\text{node}} \) is backpropagated through \( f_{\text{text}} \) to produce token-level attributions \( \mathbf{Z}^{(\hat{y})} \).

\textbf{Motivation for Token-Level Attribution.} 
As discussed in Section \ref{sec:preliminaries}, the feature matrix \( \mathbf{X} \) comprises latent text embeddings, which are challenging to interpret directly. Traditional gradient and decomposition-based methods provide node-level attributions, but they fail to attribute relevance to individual tokens in \( P \). By leveraging these methods to compute intermediate node-level attributions and extending them through relevance backpropagation, our framework produces fine-grained, interpretable token-level explanations. This approach bridges the gap between high-dimensional latent features and actionable insights. The complete steps for generating token-level attributions are detailed in Algorithm~\ref{alg:ct-lrp}, and the effectiveness of our method is demonstrated through experiments in subsequent sections.

\begin{algorithm}[t]
\caption{Contrastive Token Layerwise Relevance Propagation (CT-LRP)}
\label{alg:ct-lrp}
\DontPrintSemicolon
\SetKwInOut{}{}
\SetKwInOut{Input}{Input}\SetKwInOut{Output}{Output}
\Input{$G=(V,E)$, $P\in \mathbb{Z^+}^{|V|\times |T|}$, $f(G|\Theta)$, $f_{text}(P|\Theta_{text})$, $C$ which is set of class labels}
\Output{$\mathbf{Z}^{(\hat{y})}\in \mathbb{R}^{|V|\times |T|}$, $\textbf{Z}_{mask} = \{True, False\}^{|V|\times |T|}$}
\Begin{
encode: $f_{text}(P) = \textbf{X}$, where $\mathbf{X} \in \mathbb{R}^{|V| \times |D|}$\;
assign: $G \longleftarrow (V,E,\textbf{X})$\;
predict: $f(G) = \textbf{y}, argmax\ \textbf{y} = \hat{y}$\;
\For{$c\in C$}{
LRP for $f$: $LRP_{GNN}(f,c,G)= \textbf{R}^{(c)}$, where $\textbf{R}^{(c)} \in \mathbb{R}^{|V|\times |D|}$\;
LRP for $f_{text}$: $LRP_{text}(f_{text},\textbf{R}^{(c)},P) = \textbf{Z}^{(c)}$, where $\textbf{Z}^{(c)} \in \mathbb{R}^{|V| \times |T|}$\;
}
create: $\textbf{Z}_{mask} = \{False\}^{|V|\times |T|}$\;
\For{$c \in C$ and $c \neq \hat{y}$}{
\For{$\textbf{z}^{(c)}_{v} \in \textbf{Z}^{(c)}$}{
\For{$\textbf{z}^{(c)}_{v,t} \in \textbf{z}^{(c)}_{v}$}{
\If{$z^{(\hat{y})}_{v,t} \leq 0$}{assign: $\textbf{Z}_{mask,v,t} \longleftarrow False$}
\If{$z^{(c)}_{v,t} > 0$ and $z^{(\hat{y})}_{v,t} > 0$}{
assign: $G' = G - \textbf{x}_{t_v}$\;
do: $f(G')=\textbf{y'}$\;
\If{$\textbf{y}_{c} - \textbf{y'}_{c} > \textbf{y}_{\hat{y}} - \textbf{y'}_{\hat{y}}$}{assign: $\textbf{Z}_{mask,v,t} \longleftarrow False$}
\Else{assign: $\textbf{Z}_{mask,v,t} \longleftarrow True$}
}
\Else{assign: $\textbf{Z}_{mask,v,t} \longleftarrow True$}
}   
}
}
}
\Return $\textbf{Z}^{(\hat{y})}$, $\textbf{Z}_{mask}$
\end{algorithm}

\subsection{GNN explanation}
We select LRP~\cite{bach2015pixel} as our base method due to its strong attribution conservation properties, making it well-suited for generating interpretable explanations. Given our focus on token-level explanations, we prioritize node feature attributions while treating the structure of the input graph \( G \) as static. As a result, the importance of edges is implicitly accounted for through LRP’s attribution propagation.

For the graph convolutional and classifier layers, we employ the epsilon-stabilized LRP rule, which generates sparser explanations by focusing on the most salient features. This property is particularly valuable in large graphs, where sparsity enhances both interpretability and usability. The epsilon-stabilized LRP rule is defined as:

\begin{equation} 
    r_j = \sum_k \frac{a_j w_{jk}}{\epsilon + \sum_{j'} a_{j'} w_{j'k}} r_k 
    \label{eq:lrp-eps} 
\end{equation}

Here, \( r_j \) and \( r_k \) denote the relevance scores of neurons \( j \) and \( k \), respectively, \( a_j \) represents the activation of neuron \( j \), \( w_{jk} \) is the weight of the connection between neurons \( j \) and \( k \), and \( \epsilon \) is a small stabilizing constant. By applying this rule during the backward pass, we obtain the node feature attribution map \( \mathbf{R} \in \mathbb{R}^{|V| \times |D|} \) for the target class, with dimensions corresponding to the original node feature matrix \( \mathbf{X} \). In this attribution map, positive values indicate components that contribute to an increase in the model's output for the target class, while negative values indicate components that decrease the output for the target class. This node-level attribution serves as an intermediate representation, bridging the gap between GNN explanations and token-level attributions.

\begin{table*}[hbt!]
\caption{Average Fidelity at Fixed Sparsity for each Explanation Method applied to BiGCN, EBGCN and ClaHi-GAT trained on Twitter, Weibo and PHEME datasets. Results are obtained by taking the average of all cross-validation folds and all sparsity levels.}
\centering
\begin{tabular}{|l|c|c|c|c|c|c|}
\hline
\multirow{2}{*}{Model} & \multirow{2}{*}{Dataset} & \multicolumn{5}{c|}{Method / Fidelity}                                                                          \\ \cline{3-7} 
                       &                          & CT-LRP (Ours)             & LRP (Token)      & LRP (Node)             & c-EB             & Grad-CAM         \\ \hline
BiGCN                  & \multirow{3}{*}{Twitter} & 0.488 $\pm$0.060* & \textbf{0.526 $\pm$0.050}         & 0.081 $\pm$0.027 & 0.038 $\pm$0.008 & 0.116 $\pm$0.026 \\ \cline{1-1} \cline{3-7} 
EBGCN                  &                          & \textbf{0.674 $\pm$0.060} & 0.380 $\pm$0.013*         & 0.026 $\pm$0.024 & 0.004 $\pm$0.007 & 0.041 $\pm$0.011 \\ \cline{1-1} \cline{3-7} 
ClaHi-GAT              &                          & \textbf{0.555 $\pm$0.075} & 0.546 $\pm$0.121*         & 0.038 $\pm$0.009 & 0.000 $\pm$0.000 & 0.120 $\pm$0.022 \\ \hline
BiGCN                  & \multirow{3}{*}{Weibo}   & \textbf{0.473 $\pm$0.041} & 0.440 $\pm$0.046*         & 0.035 $\pm$0.013 & 0.001 $\pm$0.002 & 0.084 $\pm$0.005 \\ \cline{1-1} \cline{3-7} 
EBGCN                  &                          & \textbf{0.461 $\pm$0.030} & \textbf{0.461 $\pm$0.030}         & 0.036 $\pm$0.030 & 0.001 $\pm$0.001 & 0.025 $\pm$0.007 \\ \cline{1-1} \cline{3-7} 
ClaHi-GAT              &                          & \textbf{0.423 $\pm$0.047} & 0.419 $\pm$0.051*         & 0.018 $\pm$0.003 & 0.000 $\pm$0.000 & 0.105 $\pm$0.010 \\ \hline
BiGCN                  & \multirow{3}{*}{PHEME}   & \textbf{0.361 $\pm$0.063} & 0.166 $\pm$0.025*         & 0.010 $\pm$0.014 & 0.000 $\pm$0.000 & 0.003 $\pm$0.007 \\ \cline{1-1} \cline{3-7} 
EBGCN                  &                          & \textbf{0.111 $\pm$0.006} & 0.088 $\pm$0.006*         & 0.017 $\pm$0.017 & 0.002 $\pm$0.003 & 0.014 $\pm$0.020 \\ \cline{1-1} \cline{3-7} 
ClaHi-GAT              &                          & \textbf{0.482 $\pm$0.000} & 0.449 $\pm$0.012* & 0.001 $\pm$0.001 & 0.000 $\pm$0.000 & 0.001 $\pm$0.003 \\ \hline
\multicolumn{7}{l}{\footnotesize{Best result is in \textbf{Bold}. Second best result marked with (*).}} \\
\end{tabular}

\label{tab:fidelity}
\end{table*}

\begin{table*}[hbt!]
\caption{Average Sparsity for each Explanation Method applied to BiGCN, EBGCN and ClaHi-GAT trained on Twitter, Weibo and PHEME datasets. Results are obtained by taking the average of all cross-validation folds.}
\centering
\begin{tabular}{|l|c|c|c|c|c|c|}
\hline
\multirow{2}{*}{Model} & \multirow{2}{*}{Dataset} & \multicolumn{5}{c|}{Method / Sparsity}                                                                                      \\ \cline{3-7} 
                       &                          & CT-LRP (Ours)              & LRP (Token)       & LRP (Node)              & c-EB                      & Grad-CAM         \\ \hline
BiGCN                  & \multirow{3}{*}{Twitter} & 0.587 $\pm$0.049*           & {0.546 $\pm$0.011} & 0.576 $\pm$0.018  & \textbf{0.651 $\pm$0.026} & 0.205 $\pm$0.027 \\ \cline{1-1} \cline{3-7} 
EBGCN                  &                          & {0.781 $\pm$0.005*} & 0.502 $\pm$0.003           & 0.504 $\pm$0.007  & \textbf{0.568 $\pm$0.017} & 0.438 $\pm$0.055 \\ \cline{1-1} \cline{3-7} 
ClaHi-GAT              &                          & {0.743 $\pm$0.016*} & 0.492 $\pm$0.008           & 0.489 $\pm$0.008  & \textbf{1.000 $\pm$0.000} & 0.270 $\pm$0.125 \\ \hline
BiGCN                  & \multirow{3}{*}{Weibo}   & 0.585 $\pm$0.047*           & {0.502 $\pm$0.045} & 0.498 $\pm$0.062  & \textbf{0.838 $\pm$0.135} & 0.163 $\pm$0.006 \\ \cline{1-1} \cline{3-7} 
EBGCN                  &                          & {0.499 $\pm$0.012} & 0.497 $\pm$0.012           & 0.533 $\pm$0.046*  & \textbf{0.577 $\pm$0.030} & 0.413 $\pm$0.073 \\ \cline{1-1} \cline{3-7} 
ClaHi-GAT              &                          & 0.562 $\pm$0.022*           & {0.525 $\pm$0.020} & 0.545 $\pm$0.0036 & \textbf{1.000 $\pm$0.000} & 0.007 $\pm$0.009 \\ \hline
BiGCN                  & \multirow{3}{*}{PHEME}   & 0.678 $\pm$0.081*           & {0.506 $\pm$0.030} & 0.521 $\pm$0.036  & \textbf{0.917 $\pm$0.236} & 0.064 $\pm$0.078 \\ \cline{1-1} \cline{3-7} 
EBGCN                  &                          & {0.734 $\pm$0.060*} & 0.503 $\pm$0.006           & 0.503 $\pm$0.011  & \textbf{0.786 $\pm$0.218} & 0.242 $\pm$0.229 \\ \cline{1-1} \cline{3-7} 
ClaHi-GAT              &                          & 0.864 $\pm$0.068*           & {0.568 $\pm$0.077} & 0.569 $\pm$0.069  & \textbf{1.000 $\pm$0.000} & 0.467 $\pm$0.155 \\ \hline
\multicolumn{7}{l}{\footnotesize{Best result is in \textbf{Bold}. Second best result marked with (*).}} \\
\end{tabular}

\label{tab:sparsity}
\end{table*}

\subsection{Token-level explanation}


To generate token-level attributions, we backpropagate the node feature attribution map \( \mathbf{R} \) through the embedding function. The embedding process is conceptualized as a two-stage operation. In the first stage, each token \( t \) in node \( v \) is embedded into the text embedding space via the function \( f_{\text{embed}} \), producing the token vector \( \mathbf{x}_{v,t} \). In the second stage, these token vectors are aggregated to form the node vector using a pooling function \( f_{\text{pool}} \). The pooling function \( f_{\text{pool}} \) may involve simple operations like mean or max pooling, or more complex architectures such as a Multi-Layer Perceptron (MLP) or a Transformer network.



For simplicity, we first consider the case of mean pooling. The node vector for mean pooling is defined as:
\begin{equation}
    \mathbf{x}_{v,d} = \frac{1}{|T_v|} \sum_{t_v \in T_v} \mathbf{x}_{t_v,d}
    \label{eq:meanpool}
\end{equation}
where \( T_v \) is the set of tokens in node \( v \), and \( \mathbf{x}_{v,d} \) and \( \mathbf{x}_{t_v,d} \) represent the \( d^{th} \) dimension of the node vector and token vector, respectively.



Applying the epsilon-stabilized LRP rule (Eq.~\ref{eq:lrp-eps}), the relevance propagation for mean pooling is given by:
\begin{equation}
    \mathbf{r}_{t_v,d} = \frac{\mathbf{x}_{t_v,d}}{\epsilon + \mathbf{x}_{v,d}} \mathbf{r}_{v,d}
    \label{eq:lrp-meanpool}
\end{equation}




For max pooling, the node vector and its relevance propagation rule are defined as:
\begin{equation}
    \mathbf{x}_{v,d} = \max_{t_v \in T_v}(\mathbf{x}_{t_v,d})
    \label{eq:maxpool}
\end{equation}
\begin{equation}
    \mathbf{r}_{t_v,d} = 
    \begin{cases} 
        \mathbf{x}_{t_v,d} \cdot \mathbf{r}_{v,d}, & \text{if } \mathbf{x}_{t_v,d} = \max_{t_v \in T_v}(\mathbf{x}_{t_v,d}) \\
        0, & \text{otherwise}
    \end{cases}
    \label{eq:lrp-maxpool}
\end{equation}

When \( f_{\text{pool}} \) is implemented using an MLP or a more complex network, the epsilon-stabilized LRP rule (Eq.~\ref{eq:lrp-eps}) is applied to the individual layers within the network as needed.




Finally, the token-level attribution map is obtained by summing the relevance values across all dimensions of the token vectors:
\begin{equation}
    \mathbf{z}_{t_v} = \sum_{d \in D} \mathbf{r}_{t_v,d}
\end{equation}
where \( \mathbf{z}_{t_v} \) represents the total attribution of token \( t_v \) associated with node \( v \) for the target class \( c \).

\subsection{Contrastive Token-Level Explanation}
Using the framework described above, we compute token attribution maps for each class, denoted as \( \mathbf{Z}^{(c)} \) for \( c \in C \). To refine the attribution map for the predicted class \( \hat{y} \), we compare it against the attribution maps of all other classes. Specifically, we identify tokens with positive attribution values in both the predicted class and any other class. 

For each shared token, we construct a perturbed graph input \( G' = G - \mathbf{x}_{t_v} \), where the token vector \( \mathbf{x}_{t_v} \) is removed before the node feature aggregation step. The logits for the perturbed input are then calculated as \( f(G') = \mathbf{y}' \). The token's influence is determined by the difference between the original logits and the perturbed logits. If the influence on the predicted class satisfies $\mathbf{y}_{\hat{y}} - \mathbf{y}'_{\hat{y}} > \mathbf{y}_c - \mathbf{y}'_c, \quad \forall c \neq \hat{y}$, then the token contributes more strongly to the predicted class and is retained in the explanation. Otherwise, it is excluded.

To finalize the explanation, we generate a mask matrix to eliminate shared tokens that have a greater influence on other classes and tokens with negative attribution values. Multiplying the original attribution map by this mask yields an exclusive set of tokens that are specific to the predicted class, effectively disambiguating tokens that contribute positively to multiple class outputs.

An example explanation generated by CT-LRP is shown in Fig.~\ref{fig:ct-lrp-vis}. In this figure, tokens highlighted in blue and green represent class prediction-specific and general task-relevant tokens, respectively, the former exclusively contributing positively to the model's output for the predicted class and the latter contributing to multiple classes but with the most contribution to the predicted class. Tokens highlighted in red have negative attribution values for the predicted class, indicating a suppressive effect.



\section{Experiments}
\label{sec:experiments}
We validate CT-LRP through quantitative experiments on three representative GNN-based models trained on publicly available rumour detection datasets. These experiments assess the framework’s effectiveness and propose a novel paradigm for evaluating explainability in GNN-based models. This section outlines the models, datasets, and preprocessing steps, followed by the baselines and evaluation metrics. We conclude with a presentation and discussion of the experimental results.

\subsection{Models and Datasets}
\subsubsection{Models}
We evaluate CT-LRP using three representative GNN-based models, chosen for their distinct architectures and approaches to event propagation:

\begin{itemize} 
\item \textbf{Bi-Directional Graph Convolution Network (BiGCN)} \cite{Bian2020RumorDO}: This model represents event propagation as bipartite top-down and bottom-up directed graphs, processed by two separate Graph Convolution Networks (GCNs). Each GCN comprises two convolution layers, with mean pooling applied to node representations to generate graph-level embeddings. The embeddings are concatenated to form the final input for classification.

\item \textbf{Edge-enhanced Bayesian Graph Convolutional Network (EBGCN)} \cite{wei-etal-2021-towards}: Similar to BiGCN, this model uses bipartite directed graphs but introduces an edge consistency module to dynamically learn and synchronize edge weights between top-down and bottom-up GCNs. The graph aggregation strategy mirrors that of BiGCN.

\item \textbf{Claim-guided Hierarchical Graph Attention Network (ClaHi-GAT)} \cite{lin-etal-2021-rumor}: Using an undirected graph structure, this model incorporates a Graph Attention Network (GAT) with two convolution layers. It also employs source post-attention and event-level attention mechanisms to refine event representations for classification.
\end{itemize}

\subsubsection{Datasets}
We use three publicly available rumour datasets:

\begin{itemize}
\item \textbf{Twitter (combined Twitter15 and Twitter16)} \cite{ma2016detecting}: This dataset contains four-class instances, each consisting of a source post, its thread, and the source post’s label. The class labels are Non-Rumour, True-Rumour, False-Rumour and Unverified.

\item \textbf{Weibo} \cite{ma2016detecting}: This two-class dataset provides the same structure as Twitter but is based on posts from the Chinese microblogging platform Weibo.

\item \textbf{PHEME} \cite{Zubiaga2015AnalysingHP}: Similar to Twitter, this dataset also contains four-class instances. Each instance includes a source post, its thread, and a label.

\end{itemize}

For all datasets, each instance includes the original text, the source post's label, and the conversation structure, represented by the post IDs and their corresponding reply relationships. The source post has no reply relationship (null parent ID).

\subsection{Data Preprocessing and Model Training}
\subsubsection{Data Preprocessing} 
For each dataset, we use the post metadata to construct the top-down graph representing the event propagation structure. We then reverse the direction of the edges to create the bottom-up graph and combine both graphs to form the undirected version of the graph. To standardize text preprocessing across all three datasets, we use a pre-trained multilingual BERT model \cite{Devlin2019BERTPO} and its associated tokenizer to tokenize the post text and generate token embeddings. We opted for a multilingual BERT model because the PHEME dataset includes events in multiple languages. Additionally, using the same multilingual BERT model across datasets ensures that the generated explanations are consistent and not influenced by differences in text embedding models that could arise if different monolingual BERT models were used. We used mean pooling for the main experiments as they demonstrated the best overall model performance on the rumour detection task. We include the results in Appendix \ref{results}.


\subsubsection{Model Training} 
Following the task setup in the original papers for our three selected models \cite{Bian2020RumorDO, wei-etal-2021-towards, lin-etal-2021-rumor}, we train each model using a five-fold cross-validation split for the Twitter and Weibo datasets, and a nine-fold event-wise cross-validation split for the PHEME dataset. We adopt the hyperparameter settings specified in the original papers for each model and train them for 200 epochs, with early termination triggered if the loss plateaued for 10 consecutive epochs.


\subsection{Baselines and Evaluation Metrics}
\subsubsection{Baselines}
\begin{itemize} 
    \item \textbf{LRP} \cite{bach2015pixel}: A decomposition-based method that utilizes deep Taylor approximation and relevance conservation principles to explain the individual feature dimensions of nodes for the class of interest. 
    \item \textbf{Grad-CAM} \cite{Selvaraju2017ICCV}: A gradient-based method that weights the class activation map by using the mean gradient of each feature dimension of the final convolution layer for the class of interest. 
    \item \textbf{Contrastive EB (c-EB)} \cite{zhang2018top}: Another decomposition-based method that applies a probabilistic Winner-Takes-All (WTA) process to generate the attribution map for each feature dimension of the node features. 
\end{itemize}


\subsubsection{Evaluation Metrics}
We use the \textit{Fidelity} and \textit{Sparsity} metrics introduced in \cite{Pope2019ExplainabilityMF}, as well as a combined Fidelity-Sparsity metric that balances these two measures.

\begin{itemize}
    \item \textit{Fidelity} is defined as the proportion of data instances where the model's prediction changes when elements with an attribution score greater than 0.01 are removed \cite{Pope2019ExplainabilityMF}. The intuition is that removing elements deemed most salient by the model should result in a change in its prediction, indicating the importance of those elements.
    \item \textit{Sparsity} is calculated as one minus the ratio of identified elements in the explanation to the total number of nodes in the graph, i.e., $1 - \frac{m^{(c)}}{|V|}$, where $m^{(c)}$ represents the number of identified elements in the explanation for class $c$ \cite{Pope2019ExplainabilityMF}. For CT-LRP, we adapt this metric by considering the number of identified tokens in the explanation relative to the total tokens in the data instance, rather than nodes. Sparsity measures the concentration of the explanation; sparser explanations are preferred as they focus on the most salient elements, which is particularly useful in large graphs where manual inspection of every element is impractical.
    \item \textit{Fidelity-Sparsity} is the product of the \textit{Fidelity} and \textit{Sparsity} metrics. This metric captures the balance between explanation saliency and sparsity, with higher values indicating a more balanced and effective explanation.
\end{itemize}





\begin{table}[t]
\caption{Average Fidelity-Sparsity for each Explanation Method applied to BiGCN, EBGCN and ClaHi-GAT trained on Twitter, Weibo and PHEME datasets. Results are obtained by taking the product of the average Fidelity and average Sparsity.}
\centering
\begin{tabularx}{\columnwidth}{|X|X|X|X|X|X|X|}
\hline
\multirow{2}{*}{Model} & \multirow{2}{*}{Dataset} & \multicolumn{5}{c|}{Method / Sparsity}                            \\ \cline{3-7} 
                       &                          & CT-LRP (Ours)  & LRP (Token)   & LRP (Node)   & c-EB  & Grad-CAM \\ \hline
BiGCN                  & \multirow{3}{*}{Twitter} & \textbf{0.287}         & \textbf{0.287}       & 0.047 & 0.025 & 0.024    \\ \cline{1-1} \cline{3-7} 
EBGCN                  &                          & \textbf{0.527} & 0.191*               & 0.013 & 0.002 & 0.018    \\ \cline{1-1} \cline{3-7} 
ClaHi-GAT              &                          & \textbf{0.412} & 0.269*               & 0.019 & 0.000 & 0.033    \\ \hline
BiGCN                  & \multirow{3}{*}{Weibo}   & \textbf{0.277} & 0.221*               & 0.017 & 0.001 & 0.014    \\ \cline{1-1} \cline{3-7} 
EBGCN                  &                          & \textbf{0.230} & 0.083*               & 0.019 & 0.000 & 0.010    \\ \cline{1-1} \cline{3-7} 
ClaHi-GAT              &                          & \textbf{0.238} & 0.220*               & 0.010 & 0.000 & 0.021    \\ \hline
BiGCN                  & \multirow{3}{*}{PHEME}   & \textbf{0.245} & 0.084*               & 0.005 & 0.000 & 0.000    \\ \cline{1-1} \cline{3-7} 
EBGCN                  &                          & \textbf{0.081} & 0.044*               & 0.009 & 0.002 & 0.006    \\ \cline{1-1} \cline{3-7} 
ClaHi-GAT              &                          & \textbf{0.416}         & {0.255*}       & 0.000 & 0.000 & 0.001    \\ \hline
\multicolumn{7}{l}{\footnotesize{Best result is in \textbf{Bold}. Second best result marked with (*).}} \\
\end{tabularx}
\label{tab:fidelity-sparsity}
\end{table}

\begin{figure*}[!ht]
    \centering
    \includegraphics[width=\linewidth]{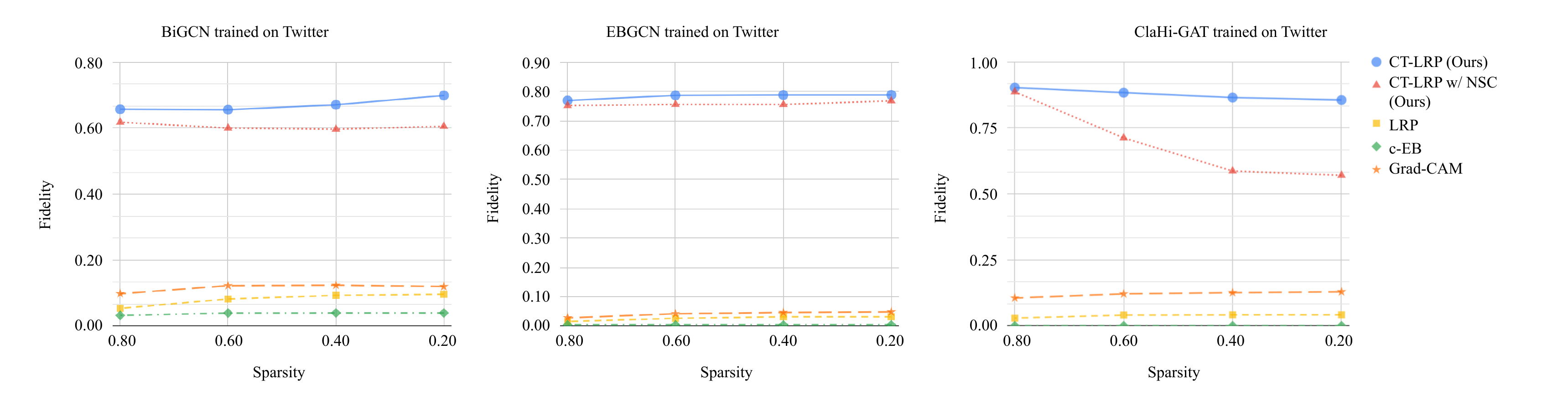}
    \includegraphics[width=\linewidth]{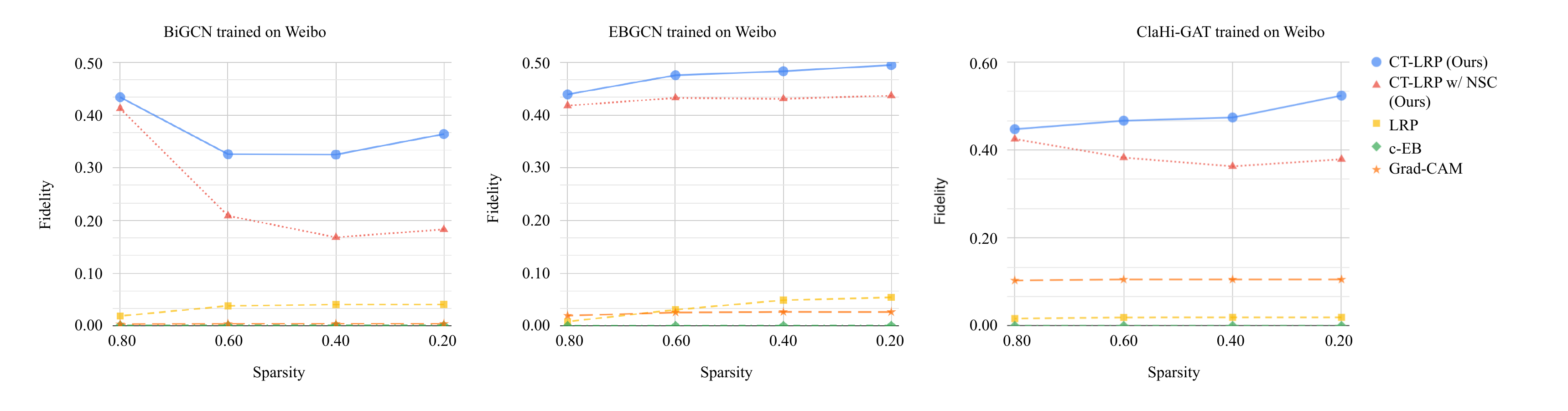}
    \includegraphics[width=\linewidth]{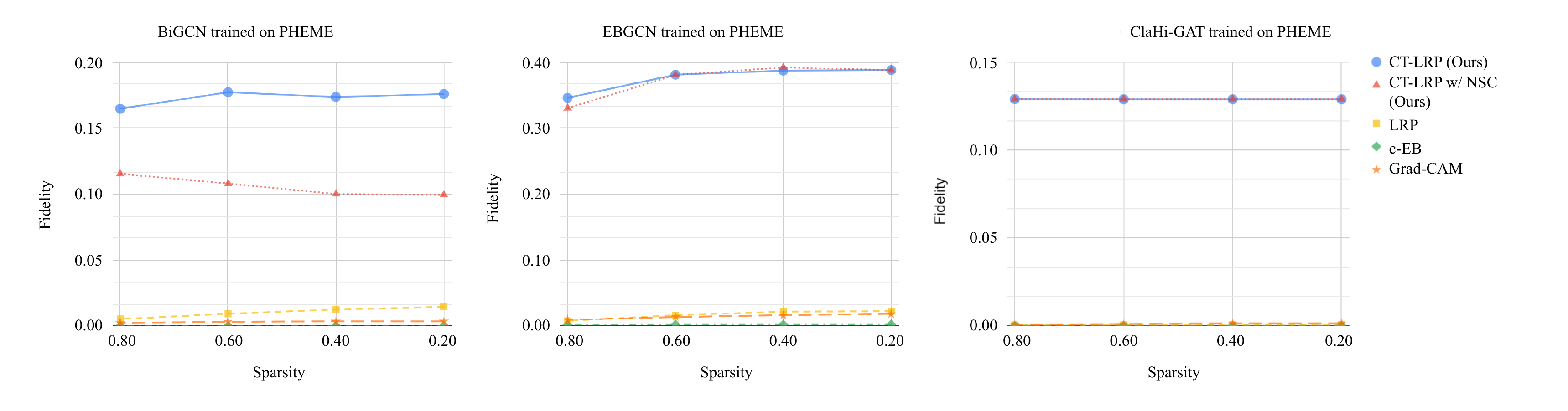}
    \caption{Quantitative performance comparisons on three models trained on three datasets with the curves obtained by varying sparsity levels.}
    \label{fig:graphs}
\end{figure*}

\subsection{Quantitative Study}
We evaluate fidelity at fixed sparsity levels for each model using k-fold cross-validation. At a given sparsity level, elements highlighted by the explanation are removed in decreasing order of importance until the sparsity limit is reached. For node-level baselines, at a sparsity of 0.8, the most salient nodes are removed until 20\% of the total nodes in the graph remain. Similarly, for CT-LRP, at the same sparsity level, the most salient tokens are removed until 20\% of the total tokens in the graph remain. To prove the effectiveness of the token class disambiguation step, we also test LRP at the token level to provide a comparison. The results are summarized in Tables \ref{tab:fidelity} and \ref{tab:sparsity}, with the top results in bold and the second-best marked with an asterisk (*).

The results show that our framework provides superior explanations of model behaviour, as evidenced by the average fidelity scores. Both CT-LRP and token-level LRP achieve the highest fidelity scores across all models and datasets. Among the two, CT-LRP performs better, showing the importance of the class membership disambiguation step, with an average of 25.77\% increase in performance over token-level LRP.




On sparsity, c-EB achieves the sparsest explanations due to its WTA process. However, its low fidelity severely limits its utility in understanding model behaviour. In contrast, the CT-LRP provide the second-highest sparsity scores while maintaining high fidelity. CT-LRP demonstrates superior performance over the baselines in achieving an optimal balance between fidelity and sparsity, as evidenced by the Fidelity-Sparsity scores in Table \ref{tab:fidelity-sparsity}, with an average 66.98\% increase in score over the next best baseline.




Fig.~\ref{fig:graphs} shows a detailed analysis of CT-LRP and other baselines at different sparsity levels for each model. Generally, as sparsity increases, CT-LRP and token-level LRP experience a notable drop in fidelity although CT-LRP exhibits a great drop in fidelity only at higher sparsities. This suggests that the disambiguation of token class membership successfully removed elements that have a greater impact on other class outputs, thereby ensuring the remaining tokens in the explanation are those which are most salient to the predicted class. On the other hand, the baselines show little change in fidelity at higher sparsity levels, likely due to the smoothing effect of node-level saliency. In such cases, removing all tokens from a node impacts the output more uniformly across classes.

\section{Discussion}
\label{sec:discussion}
CT-LRP represents a significant advancement in explainability for GNN-based rumour detection models by providing token-level explanations that are both class-specific and task-relevant. Our experiments reveal that existing GNN explainability methods, which primarily focus on node-level explanations, fall short of capturing the intricate decision-making processes underlying these models. In contrast, CT-LRP's token-level granularity offers a more nuanced understanding of model behaviour, as demonstrated by the substantial improvements in fidelity across all models and datasets. This granularity is essential for rumour detection tasks, where unseen data and dynamic content make accurate interpretation of predictions particularly challenging. 

Key findings from our application of CT-LRP to the Twitter, Weibo, and PHEME datasets underscore its ability to improve fidelity and uncover dataset-specific biases. By distinguishing between class-specific and dataset-specific tokens, our framework enables the identification of patterns and biases that might otherwise remain hidden. This insight is invaluable for researchers seeking to understand how rumour detection models generalize across datasets, as it highlights potential weaknesses and areas for improvement. Additionally, identifying biases makes these models more transparent and trustworthy, benefiting fact-checkers who rely on evidence-based explanations to substantiate their decisions.

Beyond rumour detection, CT-LRP lays the groundwork for advancing explainability in related tasks, such as fake news detection and other misinformation challenges. By framing GNN explainability as a token-level problem, it enables more trustworthy and robust AI systems. Additionally, CT-LRP is computationally efficient, requiring only a single backward pass per class, making it scalable for large graphs and real-world applications.

Despite its strengths, CT-LRP has limitations. It implicitly considers graph inductive biases during token attribution but does not explicitly explain them. Future work could address this by incorporating attribution scores that account for specific edges or subgraphs. Additionally, while our framework is tailored to a specific rumour detection task, extending it to multimodal tasks or heterogeneous graphs will require modifications to accommodate diverse data types and structures.

In summary, CT-LRP not only enhances the explainability of GNN-based rumour detection models but also sets the stage for broader applications in misinformation detection. Addressing both class-specific and dataset-specific attribution contributes to building more interpretable, trustworthy, and effective AI systems.

\section{Conclusion}
\label{sec:conclusion}
In this study, we introduced CT-LRP, a novel framework for enhancing the explainability of GNN-based rumour detection models by providing fine-grained, class-specific, and task-relevant token-level explanations. By reframing explainability from the node-level to the token-level, CT-LRP addresses critical challenges in interpreting high-dimensional latent features, achieving a new standard of granularity and fidelity in model explanations.

Our experimental results on three real-world datasets demonstrate the framework's effectiveness, highlighting its superiority in generating interpretable and high-fidelity explanations compared to existing methods. Furthermore, CT-LRP's ability to identify both class-specific and dataset-specific patterns sheds light on potential biases and generalization gaps, offering valuable insights for improving model reliability and fairness.

Beyond rumour detection, the principles underlying CT-LRP extend to broader misinformation detection and graph-based applications, making it a versatile tool for advancing trust in AI systems. As misinformation continues to pose significant societal risks, CT-LRP represents a meaningful step toward building transparent, trustworthy, and ethical AI solutions. Future work will focus on extending this framework to multimodal and heterogeneous graphs, exploring its potential in increasingly complex real-world scenarios.



\appendices
\label{sec:appendix}
\section{}
\label{results}
\begin{table}[h!]
\caption{Model Rumour Detection Accuracy with Different Text Embedding Pooling Functions}
\centering
\label{tab:prediction}
\begin{tabular}{|l|l|c|c|c|}
\hline
\multirow{2}{*}{Model}     & \multirow{2}{*}{Pooling Method} & \multicolumn{3}{c|}{Dataset}                      \\
\cline{3-5}
                           &                                 & Twitter        & PHEME          & Weibo          \\
\hline
\multirow{3}{*}{BiGCN}     & Mean                            & \textbf{0.750} & 0.416          & \textbf{0.933} \\
                           & Max                             & 0.659          & \textbf{0.579} & 0.888          \\
                           & CLS                             & 0.575          & 0.521          & 0.845          \\
\hline
\multirow{3}{*}{EBGCN}     & Mean                            & \textbf{0.756} & 0.423          & \textbf{0.928} \\
                           & Max                             & 0.746          & 0.409          & 0.911          \\
                           & CLS                             & 0.648          & \textbf{0.566} & 0.905          \\
\hline
\multirow{3}{*}{ClaHi-GAT} & Mean                            & \textbf{0.677} & 0.423          & \textbf{0.929} \\
                           & Max                             & 0.652          & \textbf{0.484} & 0.916          \\
                           & CLS                             & 0.561          & 0.426          & 0.879         \\
\hline
\multicolumn{5}{l}{\footnotesize{Best result is in \textbf{Bold}.}} \\
\end{tabular}
\end{table}

\bibliographystyle{IEEEtran}
\bibliography{IEEEabrv, bib}

\begin{IEEEbiography}[{\includegraphics[width=1in,height=1.25in,keepaspectratio]{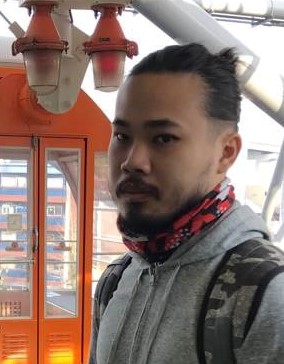}}]{Daniel Wai Kit Chin} completed his Bachelor's degree in Engineering from Singapore University of Technology and Design in 2019. He is currently a PhD student in the Information Systems Technology and Design Pillar, Singapore University of Technology and Design. His research interests are in Explainable Artificial Intelligence, Graph Neural Networks and Social Computing.
\end{IEEEbiography}
\vskip -2\baselineskip plus -1fil
\begin{IEEEbiography}
[{\includegraphics[width=1in,height=1.25in,keepaspectratio]{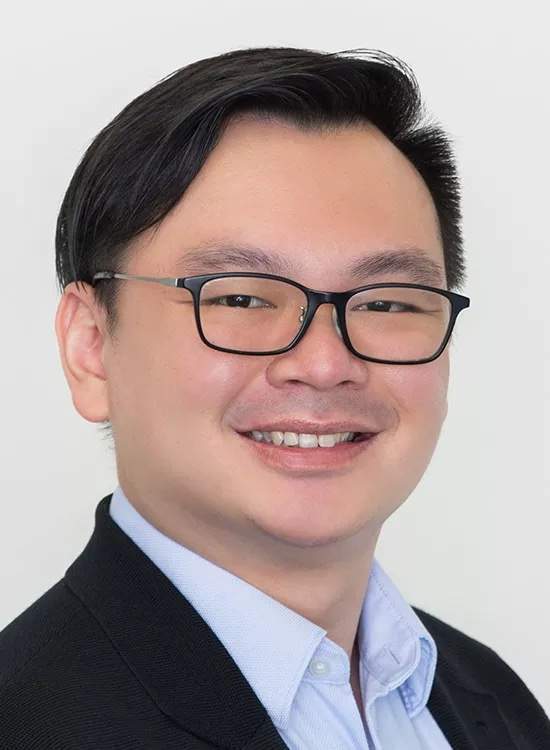}}]{Roy Ka-Wei Lee} is an Assistant Professor at the Information Systems Technology and Design Pillar, Singapore University of Technology and Design. He is a faculty of the transformative Design and Artificial Intelligence programme. His research lies in the intersection of data mining, machine learning, social computing, and natural language processing. He is leading the Social AI Studio, a research group that focuses on designing next-generation social artificial intelligence systems. He has published in top-tier venues in data mining and computation linguistics domains. He serves on the program committees of multiple top data mining and natural language processing conferences.
\end{IEEEbiography}
\end{document}